\documentclass[letterpaper, 10 pt, conference]{ieeeconf}  

\IEEEoverridecommandlockouts                              

\usepackage{times}

\usepackage{multicol}
\usepackage[dvipsnames]{xcolor}
\usepackage{graphicx}
\usepackage{hyperref}
\usepackage{amsmath}
\usepackage{cleveref}
\usepackage[ruled,vlined,linesnumbered]{algorithm2e}
\usepackage[font=footnotesize,labelfont=bf]{caption}
\usepackage{subcaption}
\usepackage{bbm}
\usepackage{float}
\usepackage{color, soul} 
\usepackage{wrapfig,lipsum,booktabs}

\usepackage{array}
\usepackage[utf8]{inputenc}
\usepackage{amsmath, amssymb}
\usepackage{booktabs}
\usepackage{dsfont}
\usepackage{diagbox}
\usepackage{multirow}

\usepackage{amsmath,amsfonts,bm}

\newcommand{\newterm}[1]{{\bf #1}}

\def\eqref#1{equation~\ref{#1}}

\def\1{\bm{1}}

\def\va{{\bm{a}}}

\def\vr{{\bm{r}}}
\def\vs{{\bm{s}}}

\def\vx{{\bm{x}}}

\def\vz{{\bm{z}}}

\def\mE{{\bm{E}}}

\def\mN{{\bm{N}}}

\def\mP{{\bm{P}}}

\def\mS{{\bm{S}}}

\DeclareMathAlphabet{\mathsfit}{\encodingdefault}{\sfdefault}{m}{sl}
\SetMathAlphabet{\mathsfit}{bold}{\encodingdefault}{\sfdefault}{bx}{n}

\def\gD{{\mathcal{D}}}

\def\gL{{\mathcal{L}}}

\newcommand{\KL}{D_{\mathrm{KL}}}

\newcommand{\btau}{\boldsymbol{\tau}}

\newcommand{\br}{\mathbf{r}}

\title{\LARGE \bf

Implicit Contact Diffuser: Sequential Contact Reasoning \\
with Latent Point Cloud Diffusion

}

\author{Zixuan Huang$^{1}$, Yinong He$^{*1}$, Yating Lin$^{*1}$, Dmitry Berenson$^{1}$
\thanks{$^{1}$ University of Michigan, Ann Arbor}
\thanks{$^*$ equal contribution }
}

\begin{document}
\newcommand{\scenepc}{\mP_s}
\newcommand{\objpc}{\mP_o}
\newcommand{\ndfpc}{\mP_{ndf}}
\newcommand{\ndfpci}[1]{\mP_{ndf_{#1}}}
\newcommand{\method}{\textit{Implicit Contact Diffuser}\xspace}
\newcommand{\methodshort}{\text{ICD}\xspace}



\maketitle
\thispagestyle{empty}
\pagestyle{empty}

\begin{abstract}
Long-horizon contact-rich manipulation has long been a challenging problem, as it requires reasoning over both discrete contact modes and continuous object motion. We introduce \method~(\methodshort), a diffusion-based model that generates a sequence of neural descriptors that specify a series of contact relationships between the object and the environment. This sequence is then used as guidance for an MPC method to accomplish a given task. The key advantage of this approach is that the latent descriptors provide more task-relevant guidance to MPC, helping to avoid local minima for contact-rich manipulation tasks. 
Our experiments demonstrate that \methodshort outperforms baselines on complex, long-horizon, contact-rich manipulation tasks, such as cable routing and notebook folding. 
Additionally, our experiments also indicate that \methodshort can generalize a target contact relationship to a different environment. More visualizations can be found on our website \href{https://implicit-contact-diffuser.github.io/}{https://implicit-contact-diffuser.github.io}

\end{abstract}

\section{Introduction}
\begin{figure}[t]
    \vspace{2mm}
    \centering
    \includegraphics[width=0.48\textwidth]{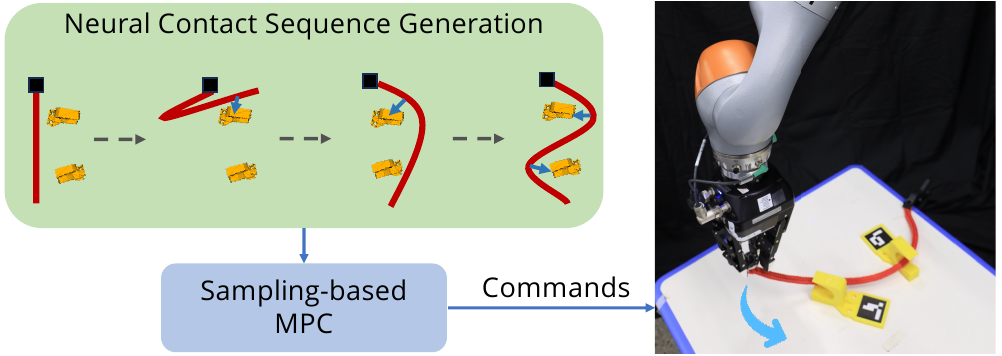}
    \caption{By predicting future contact sequences using a latent diffusion model, we enable long-horizon contact-rich deformable object manipulation such as cable routing using a sampling-based MPC controller.}
    \label{fig:teaser}
    \vspace{-5mm}
\end{figure}

Interacting with the environment through contact is central to many robotic tasks, such as manipulation and locomotion. Despite the ubiquity of contact interactions, controlling these hybrid systems poses significant challenges due to the complex interplay between discrete contact events and continuous motion. For instance, in cable routing, the robot must generate smooth motions to initiate and maintain contact between the cable and the fixtures (Fig.~\ref{fig:teaser}). If the contact breaks at any point, the cable could slip off the fixtures. Moreover, when model errors or external disturbances occur, the robot must adjust its actions accordingly to maintain task success.

A large body of work has attempted to tackle these challenges by planning~\cite{cheng2021contact, cheng2022contact, cheng2023enhancing, aceituno2022hierarchical} or trajectory optimization~\cite{park2007convex,mordatch2012contact,posa2014direct} through contact. However, these methods are typically limited to rigid objects, or face limitations in online replanning due to the high computational costs involved.

In this paper, we introduce a learning-based model predictive control (MPC) framework to address this class of problems. In particular, we train a latent diffusion model to generate future contact sequences as subgoals, which guide a MPC controller to generate robot motions that establish the desired contact relationships. A key question, however, is determining the best representation for these contact relationships.

One approach is to use binary contact states. Wi et al.~\cite{wi2023calamari} propose to specify desired contact locations by predicting a heatmap over the environment. However, this approach lacks critical information regarding which part of the object should be in contact, a crucial factor for tasks where maintaining precise object-environment interactions is important. Additionally, it cannot capture the dynamic contact switching required in certain tasks.

To overcome these limitations, we leverage recent advancements in implicit neural representations and encode contact relationships using a modified version of Neural Descriptor Fields (NDF)~\cite{simeonov2022neural}. We train a scene-level NDF to capture geometric information by predicting occupancy and gradient direction of the signed distance function. By querying the scene NDF with the object’s point cloud, we compute a dense, contact-aware representation of the object. Our experiments show that these neural descriptors capture task-relevant geometric relationships (e.g., left or right of a fixture) rather than specific locations, providing more flexible guidance. This allows us to transfer goal contact relationships across different environments at test time.

To capture the contact switching required to reach a goal, we train a latent diffusion model to predict the contact sequence represented by neural descriptors. We also learn a reachability function, similar to Subgoal Diffuser~\cite{huang2024subgoal}, to determine the required sequence length. The key contributions of this paper are: 1) a latent diffusion model that reasons about evolving contact relationships in long-horizon manipulation tasks; 2) an MPC framework that plans motions based on desired contact relationships rather than precise locations. 3) a scene-level neural descriptor field that provides local contact representations, enabling greater generalization across environments.

We validate our method on challenging long-horizon contact-rich manipulation tasks, including cable routing and notebook folding.  Our results show that \methodshort outperforms or is on par with baselines that plan to exact locations rather than focusing on contact relationships, as well as baselines that directly predict actions without planning. \methodshort can also adapt a target contact relationship to a different environment naturally.


\section{Related work}
\subsection{3D Representation for Object Manipulation}
Prior works studies different representations for object manipulations, such as key points~\cite{manuelli2019kpam}, RGB image~\cite{hoque2022visuospatial, chi2023diffusion}, point cloud~\cite{yang2023equivact, ze20243d} or mesh~\cite{lin2022learning, huang2022mesh, huang2023self}.
Recently, Neural Descriptor Fields (NDF)~\cite{simeonov2022neural,simeonov2023se, chun2023local} demonstrates itself as an effective implicit representation for category-level generalization. In this work, we propose a variant of NDF where spatial structure is preserved. We show that compared to explicit representations such as point cloud, the NDF better captures the soft contact relationships between object and environment.

\subsection{Contact Reasoning for Robot Manipulation}
Controlling the robot to make and break contacts purposefully has been one of the key challenges for robotics, since it involves optimizing over a hybrid system that contains both continuous (robot motion) and discrete variables (contact). One common approach~\cite{cheng2021contact, cheng2022contact, cheng2023enhancing, aceituno2022hierarchical} is to find object motions using a sampling-based motion planner guided by a high-level search for contact modes. However, these methods are typically limited to rigid objects.
Recently, learning-based methods have been introduced to detect or control contact~\cite{wi2022virdo, wi2022virdo++,van2023learning, van2023integrated, higuera2023neural} for complaint tools such as spatulas.
Wi et al.~\cite{wi2023calamari} designs a framework for contact-rich manipulation that predicts the target contact patch over the environment conditioned on the language. However, the predicted contact patch does not specify which part of the object should make contact, and does not model a sequence of changing contacts.
In this paper, we propose to use a contact-aware neural representation and a diffusion-based architecture to model future contact sequences for highly deformable objects, such as cables.

\subsection{Diffusion Models for Robotics}
Diffusion models have also been applied to robot manipulation, either as a policy class that predicts action directly from observation~\cite{chi2023diffusion, gervet2023act3d, ha2023scaling, chen2023playfusion, shi2023waypoint, mishra2023generative, scheikl2024movement}, or as a learned planner to generate future trajectories~\cite{janner2022planning, kapelyukh2023dall, ajay2022conditional, li2023hierarchical, huang2024subgoal}. 
Although some existing diffusion-based methods have been shown to work on certain contact-rich manipulation tasks, such as planar pushing~\cite{chi2023diffusion}, dumpling making~\cite{ze20243d} or book shelving~\cite{simeonov2023shelving}, our experiment suggests that they struggle with tasks that involve long-horizon reasoning of changing contacts.
Similarly to us, the Subgoal Diffuser~\cite{huang2024subgoal} generates future subgoals using a diffusion model to guide an MPC controller. However, Subgoal Diffuser represents the subgoals using locations of key points, which can be overly constrained and does not reason about the contact interaction between object and environments explicitly.

\subsection{Long-horizon reasoning for robot manipulation}
Long-horizon manipulation tasks usually contain several distinct stages and contain a lot of local optima. One way to tackle this is to plan over skill abstractions~\cite{eysenbach2019search, shi2023robocook, lin2022diffskill, lin2022planning, cheng2023league, jin2022robotic, luo2024multi} learned with imitation learning or reinforcement learning. Another way is to decompose tasks into multiple subgoals~\cite{jurgenson2020sub,nair2019hierarchical,fang2022planning, nasiriany2019planning, huang2024subgoal, li2023hierarchical}, which can be used to guide a low-level policy.
We propose a method to generate subgoals represented by neural descriptors, which will highlight the contact relationships bwtween the objects and environment. While NOD-TAMP~\cite{cheng2023nod} uses a similar representation for long-horizon reasoning, it adapts a given demonstration trajectory to a new situation by optimization while we directly learn the distribution of the trajectory using a latent diffusion model. Also, NOD-TAMP cannot handle deformable objects.

\begin{figure*}[t]
    \centering
    \includegraphics[width=\linewidth]{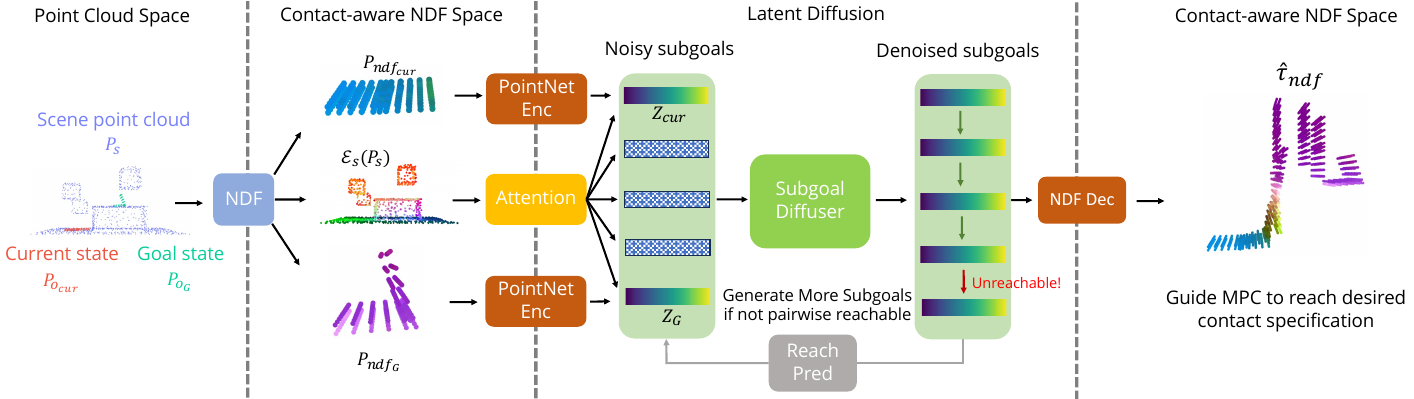}
    \caption{
    System overview, with the notebook folding task as an example. First, \methodshort transforms the scene, current object, and goal object point cloud, into an implicit contact representation using a modified NDF model. The NDF model can be used to extract point-wise contact relationships of the object, shown by the color. Next, we project the dense NDF point clouds into low-dimensional latent vectors and utilize a latent diffusion model to generate a sequence of contact subgoals. The latent diffusion model generates subgoals recursively from coarse to fine, depending on a reachability measure. Finally, we track these predicted subgoals using a sampling-based MPC method, ensuring that the object reaches the desired contact specification.
    }
    \label{fig:sys1}
    \vspace{-5mm}
\end{figure*}

\section{Preliminaries}
\subsection{Problem Statement}
\label{sec:problem_statement}
In this paper, we consider long-horizon contact-rich manipulation problems of deformable object that involve changing contacts. 
We denote the robot state by $\vs_t$ and the action by $\va_t$. The goal specification is represented as a pair of point clouds $(\mP_{o_g}, \scenepc)$, where $\mP_{o_g}$ is the point cloud of the object in a goal state and $\scenepc$ is the point cloud of the scene.
However, the goal is not to match the shape and pose of the object exactly but to match the \textbf{contact relationship} between the object and the scene, so that the object is in contact with the scene in the correct locations. For example, in a cable routing task, the objective is to route the cable through the opening of the hook, ensuring that the cable touches the front side of the hook but not the back. It is important to note that we focus solely on the geometric aspect of the contact, without differentiating between the contact modes such as sticking or sliding contact.

This type of problem presents significant challenges due to the need for joint reasoning over both continuous motion and discrete contact switching, particularly for high-dimensional deformable objects. Additionally, long-horizon reasoning is crucial for generating effective contact-switching behavior while avoiding local minima, ensuring that the robot can progress toward the final goal without becoming stuck in suboptimal configurations.

Our objective is to learn a dense object-centric representation of contact relationship, which describes how each point of the object interacts with the environment. Next, we learn a generative model that, given the current state, scene, and goal specification, predicts a sequence of contact subgoals using the learned representation.
These subgoals guide an MPC method to sequentially make and break contact and ultimately reach a goal state that conforms to the goal specification. 


We assume access to an offline dataset $\gD$, which contains N different trajectories of object point clouds and the corresponding scene point cloud $(\btau^i, \scenepc^i)$, where $\btau^i = [P_{o_0}^i, P_{o_1}^i, \cdots, P_{o_L}^i]$. The offline dataset is collected with a scripted policy that does not guarantee task completion. 
We also assume access to the full point cloud for both the object and the scene, and the order of points in $\objpc$ does not change between states.

\subsection{Diffusion Models}
\method is largely based on diffusion models~\cite{sohl2015deep, ho2020denoising}, which are a powerful class of generative models that frame data generation as a $K$-step iterative denoising procedure. To sample a noise-free output $\btau*^0$ from a diffusion model, it starts by sampling $\btau^K$ from a Gaussian noise distribution. Then we perform $K$ iterations of stochastic Langevin Dynamics~\cite{welling2011bayesian} with the update rule $   \btau^{k-1} = \alpha^k(\btau^{k} - \gamma^k \epsilon_\theta (\btau^{k},k) + N(0, \sigma^2I))$.
 $\alpha^k$ and $\gamma^k$ are both hyperparameters related to the noise schedule and $N(0, \sigma^2I))$ denotes Gaussian noise added at each iteration. 
 $\epsilon_\theta$ is parameterized by a neural network to estimate the noise that can be used to recover the original data. DDPMs~\cite{ho2020denoising} propose to train diffusion model using the variational lower-bound on $\log p_\theta(\btau)$: $ \gL_{DDPM}(\theta)=||\epsilon^k - \epsilon_\theta(\btau^k, k)||^2$.

\section{Method}
In this section, we introduce \method, a method designed to capture and reason about contact switching in long-horizon deformable object manipulation.  In Section \ref{method:descriptor}, we discuss how to represent the object-environment contact relationships of deformable objects using an implicit neural representation. In Section \ref{method:latent_diffusion}, we describe how to train a latent point cloud diffusion model to predict the contact sequence. Finally, in Section~\ref{method:mpc}, we discuss how to follow the predicted contact sequence using a sampling-based MPC planner.

\subsection{Contact-aware Neural Descriptor Field}
\label{method:descriptor}
Finding a suitable contact representation that facilitates planning is a challenging problem. If we naively represent contact with a binary discrete representation, planning over the contact space can quickly become combinatorially expensive, which is one of the reasons why prior methods~\cite{xian2023chaineddiffuser,aceituno2022hierarchical} struggle with deformable objects.
Our key insight is that we can capture the soft object-environment contact relationships using a continuous implicit neural representation. We build upon Neural Descriptor Fields (NDF)~\cite{simeonov2022neural,simeonov2023se,chun2023local} to develop a contact-aware neural representation for deformable objects, utilizing a scene NDF. Given a scene point cloud $\scenepc$, we learn a function $f$ to map a 3D coordinate $x \in \mathbb{R}^3$ to a latent neural descriptor in $\mathbb{R}^d$:
\begin{equation}
    f(\vx|\scenepc) = f(\vx|\mathcal{E}_s(\scenepc))
\end{equation}
\noindent where $\mathcal{E}_s(\scenepc)$ is a PointNet~\cite{qi2017pointnet} model. 
Given an object point cloud $\objpc$, the state of the object can be described as the concatenation of all point descriptors:
\begin{equation}
   \ndfpc = \phi_{NDF}(\objpc|\scenepc)= \bigoplus_{\vx_i\in \objpc} f(\vx_i|\scenepc)
\end{equation} 
Since the function $f$ is trained to predict the geometric features of the scene, the NDF point cloud $\ndfpc\in \mathbb{R}^{N\times d}$ can be interpreted as an encoding of point-wise geometric relations with the scene for every point on the object.

\begin{figure*}[t]
    \centering
    \includegraphics[width=0.9\linewidth]{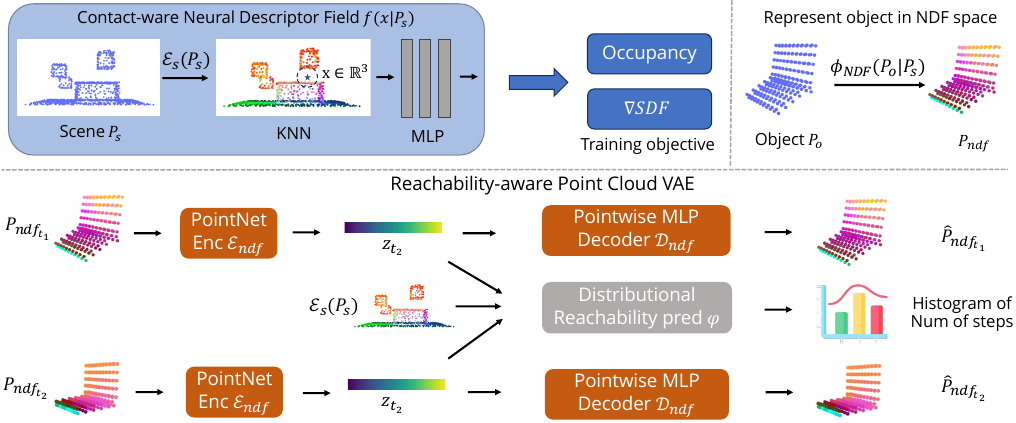}
    \caption{
    As shown in the the upper figure, the NDF model is trained to encode local geometries of the scene by predicting occupancy and gradient direction of the Signed Distance Function (SDF) of the scene. Given an object point cloud $\objpc$, such as that of a notebook, we transform it into a contact-aware latent representation $\ndfpc$.
    In the bottom figure, we show how the reachability-aware point cloud VAE is trained. In additional to the regular reconstruction and KL divergence loss, we introduce a distributional reachability prediction loss to encourage temporal consistency in the latent space. The reachability predictor is also used in the latent diffusion model to decide the number of subgoals required for the tasks, as shown in Fig.~\ref{fig:sys1}.
    }
    \label{fig:ndf_training}
    \vspace{-5mm}
\end{figure*}
We make several key design choices to adapt NDF, ensuring it better suits the tasks we are dealing with.
Similar to Simeonov et al.~\cite{simeonov2022neural}, we train $f(\vx|\scenepc)$ using occupancy prediction. Additionally, we incorporate an auxiliary loss on the gradient direction of the signed distance function (SDF): $\mathcal{J}_{grad} =(\nabla SDF(x)-\hat{\nabla} SDF(x))^2$, where $\nabla SDF(x)$ and $\hat{\nabla} SDF(x))$ refer to ground-truth and predicted gradients of the SDF. 
This helps the descriptors encode not only whether a point is in contact (occupied), but also how to make contact for points that are not yet in contact.

NDF adopts a $\mS \mE(3)$-invariant neural network architecture, Vector Neuron~\cite{deng2021vector}, to enhance the generalizability of the descriptors. While the descriptors remain unchanged when a transformation $T\in \mS \mE(3)$ is applied to the object and the scene simultaneously, this can sometimes lead to unrealistic outcomes. For example, the object will have similar NDF features whether it contacts the floor or the ceiling. To mitigate this, we modify the Vector Neuron to be invariant only to the rotations along the direction of gravity (as gravity plays a large part in determining the configuration of a deformable object) , which we define as $\mS \mE(3)^z$. Specifically, we add a small constant value to the z-axis of the point features, ensuring that rotations not aligned with the z-axis produce distinct latent features. 


The original NDF model~\cite{simeonov2022neural} encodes the entire point cloud into a single global feature vector by averaging over $\mathcal{E}_s(\scenepc)$. In contrast, we aggregate the local features of nearby contact candidates for each query point using K-nearest neighbors (Fig.~\ref{fig:sys1}) based on the intuition that the object is more likely to make contact with spatially closer points. Our experiments indicate that incorporating these local NDF features is important for improving task performance.

\subsection{\method}
\label{method:latent_diffusion}
In the previous section, we describe a dense contact-aware neural representation for deformable objects. Now we will use this representation to tackle long-horizon contact-rich manipulation problems with contact switching. We introduce \newterm{Implicit Contact Diffuser}, a diffusion-based architecture that generates a sequence of subgoals $\btau_{ndf} =[\ndfpci{0}, \ndfpci{1}, \ndfpci{2}, \dots , \ndfpci{M}]$, represented as NDF point clouds $\ndfpc \in \mathbb{R}^{N\times d}$. 

While diffusion models have been applied to point cloud generation, prior works~\cite{luo2021diffusion, vahdat2022lion} only generate individual point clouds $\mP\in\mathbb{R}^{N \times 3}$. In our case, in order to capture contact switching, we need to generate a sequence of coherent latent point clouds consisting of high-dimensional point features.

To tackle this sequential point cloud generation problem, we propose using Latent Diffusion Models (LDM)~\cite{rombach2022high}. We begin by training a Variational Autoencoder (VAE)~\cite{kingma2013auto} to project the high-dimensional point cloud $\ndfpc$ into low-dimensional vectors. Next, we train a hierarchical diffusion model to recursively generate subgoals from coarse to fine, following Huang et al.~\cite{huang2024subgoal}.

\noindent\textbf{Reachability-aware Point Cloud VAE}. The VAE comprises three components: a PointNet++ encoder $\mathcal{E}_{ndf}(\vz_t|\ndfpci{t})$ \cite{qi2017pointnet++}, a point-wise MLP decoder $\mathcal{D}_{ndf}(\hat{\mP}_{ndf_t}| \objpc^{canon}, \vz_t)$, and a distributional reachability prediction MLP $\varphi(\hat{\vr}|\vz_{t_{1}}, \vz_{t_{2}}, \mathcal{E}_s(\scenepc))$, as visualized in Fig.~\ref{fig:ndf_training}.
The encoder $\mathcal{E}_{ndf}$ compresses the NDF point cloud $\ndfpci{t}$ into a latent vector $\vz_t$. The pointwise MLP decoder $\mathcal{D}_{ndf}$ is adapted from Luo et al.~\cite{luo2021diffusion}. Given $\vz_t$ and the canonical object point cloud $\mP_o^{c}$, an implicit decoder $\mathcal{D}_{ndf}$ reconstructs the NDF point cloud from the latent vector. The query coordinates $\mP_o^{canon}$ are predefined, i.e., a straight rope or a magazine that is laid flat. 

The VAE is trained by three different losses:

\begin{center}
    \vspace{-6mm}
    \begin{align}
    \mathcal{L}_{vae} &= \lambda_1\mathcal{L}_{recon}(\ndfpci,\hat{\mP}_{ndf}) \\
    &+ \lambda_2\KL(\mathcal{E}_{ndf}(\vz_t|\ndfpci{t}), \mN(z))  \\
    &+ \lambda_3\mathcal{L}_{Reach}(\vr, \varphi(\vr|\vz_{t_{1}}, \vz_{t_{2}},\mathcal{E}_s(\scenepc)))
\end{align}
\end{center}

In addition to the regular reconstruction loss and KL regularization loss, we introduce a reachability loss $\mathcal{L}_{reach}$ to encourage temporal consistency in the learned latent space.

During training, we sample pairs of states in the same trajectory using the discounted state occupancy measure (lower probability for states that take more steps to reach) in line with previous work~\cite{eysenbach2024inference, eysenbach2022contrastive}. For a pair of NDF point clouds $(\ndfpci{t_{1}}, \ndfpci{t_{2}})$, we define reachability as the minimum number of steps to travel between them. Following Subgoal Diffuser~\cite{huang2024subgoal}, we discretize the reachability into $K$ bins and frame the reachability prediction problem as a classification problem and train an MLP $\varphi(\vr|\vz_{t_{1}}, \vz_{t_{2}}, \mathcal{E}_s(\scenepc))$ with cross-entropy loss.
Since we do not assume that the training data are high-quality demonstrations, and there might exist multiple paths of different lengths to travel between two states,  $\varphi(\vr| \vz_{t_{1}}, \vz_{t_{2}}, \mathcal{E}_s(\scenepc)$ will capture the distribution of reachability between two states.
During test time, we use ``softmin'' to estimate shortest distance (highest reachability), which is used to determine the number of subgoals for the latent diffusion model.

\noindent\textbf{Latent Point Cloud Diffusion Model} 
\label{sec:method:latent_diffusion}
The objective of the latent diffusion model is to generate a sqeuence of NDF subgoals $\btau_{ndf}$,
given current state, goal specification, and the scene. With the point cloud VAE described above, the diffusion model only needs to model the distribution of the condensed latent vectors, denoted as $p(\btau_{z} | \vz_{cur}, \vz_{goal}, \mathcal{E}_s(\scenepc))$. Following Subgoal Diffuser~\cite{huang2024subgoal}, we generate subgoal sequences recursively in a coarse to fine manner. Starting from $\btau_z^0=[\vz_{cur}, \vz_{goal}]$, in each iteration, the number of subgoals in $\btau_z^{l+1}$ increases by 
$|\btau_z^{l+1}|= |\btau_z^{l}|\times 2 -1$. Instead of generating from scratch, the latent diffusion model predicts the next level of subgoals $\btau_z^{l+1}$ conditioned on the previous ones $\btau_z^{l}$. Hence, the latent diffusion model can be written as $p(\btau_z^{l+1}|\btau_z^l, \mathcal{E}_s(\scenepc))$.

\subsection{MPPI with Neural Contact Subgoals}
\label{method:mpc}
Every $T$ steps, Neural Contact Diffuser generates a sequence of contact subgoals $\hat{\btau}_{ndf}$.
We use a sampling-based MPC method, Model Predictive Path Integral (MPPI)~\cite{williams2016aggressive}, to plan a sequence of robot actions to track the subgoals.
We define robot actions $\va \in \mathbb{R}^3$ as the delta translation of the end effector. 
At each step, MPPI samples $K$ action sequences of length $H$, where $H$ is the planning horizon.

The sampled actions are evaluated by rolling out in the MuJoCo~\cite{todorov2012mujoco} simulator with the following cost:
\begin{align*}
   \mathcal{J}_{MPPI} =  \sum_{t=0}^{H-1} &
    \Biggl( \min_{\hat{\mP}_{ndf_i} \in \hat{\btau}_{ndf}} (\mP_{ndf_t}-\hat{\mP}_{ndf_i})^2 \\ 
        &+\lambda_{col}\max(-SDF(\br_t),0) \Biggr)
\end{align*}
$\hat{\btau}_{ndf}$ is the desired NDF subgoals predicted by the diffusion model. The rollouts from the simulator are transformed to NDF space using $\phi_{ndf}$, which we denote as $\mP_{ndf_t}$. The first cost term is  the Euclidean distance to the closest NDF subgoal. A subgoal will be removed from the goal chain $\hat{\btau}_{ndf}$ once the current state is within predefined distance threshold. The second cost is to prevent the robot from colliding with the environment, represented as the scene SDF. The robot geometry is approximated by a set of spheres as in ~\cite{sundaralingam2023curobo}.
By minimizing $\mathcal{J}_{MPPI}$, MPPI generates robot actions that manipulate the object to achieve the desired contact relationships described by $\hat{\btau}_{ndf}$.

\section{Experiments}
Our experiments aim to show that 1) the scene NDF is a good representation for capturing contact relationships and 2) \method\ is capable of long-horizon contact reasoning and generating contact sequences to guide an MPC controller to reach the desired contact relationship. We also demonstrate our method on a physical robot and the videos can be found on our \href{https://implicit-contact-diffuser.github.io/}{website}.

\begin{figure}[t]
    \centering
    \includegraphics[width=0.4\textwidth]{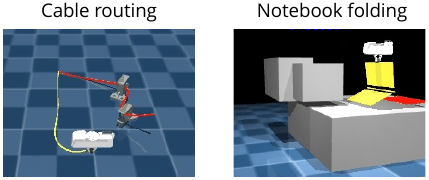}
    \caption{We evaluate our methods on two long-horizon contact-rich tasks in simulation: cable routing and notebook folding. Goals are visualized in red.}
    \vspace{-3mm}
    \label{fig:sim_tasks}
\end{figure}

\begin{table}
    \centering
    \begin{tabular}{@{}c|cc|c@{}}
        \toprule
        Method & \multicolumn{2}{c|}{Cable Routing} & Notebook \\
               & Success $\uparrow$ & Complete $\uparrow$ & Success $\uparrow$  \\
        \midrule
        \textbf{\method}                               & \textbf{90} & \textbf{95} & 95 \\
        Subgoal Diffuser~\cite{huang2024subgoal}       & 65  & 80 & \textbf{100} \\
        Diffusion Policy~\cite{chi2023diffusion}       & 30  & 40 & 70 \\
        3D Diffusion Policy~\cite{ze20243d}            & 15  & 40 &  5 \\
        PC-MPPI                                        & 25  & 55 &  50\\
        NDF-MPPI                                      & 55  & 70 & 10 \\
        \midrule
        Global NDF                                     &50 &75 & 75\\ 
        \hline
    \end{tabular}
\caption{We evaluate every method on 10 test cases for 2 seeds (20 runs in total) and report the success rate. For the cable routing task, success is defined as the cable being routed through both fixtures. Additionally, we report the "complete rate," which represents the percentage of fixtures successfully routed by the cable.}
\label{table:sim_result}
\vspace{-6mm}
\end{table}
\subsection{Simulation Experiments}
\subsubsection{Tasks}
We evaluate our method on two long-horizon manipulation tasks that involve changing contact (Fig.~\ref{fig:sim_tasks}). 

\noindent\textbf{Cable routing}. The goal is to route the rope through two randomly placed fixtures on a table. One end of the cable is fixed and the other is grasped by a floating gripper. The task is considered successful if the rope is routed through \emph{both} fixtures. We also consider the ``complete rate''---the percentage of successfully routed individual fixtures. This task is challenging due to: 1) The high-dimensional state space and complex rope dynamics; 2) The need for precise control of a deformable object (ensuring the cable stays inside the first fixture when routing the second); and 3) Long-horizon reasoning (to avoid local minima).

\noindent\textbf{Notebook folding}. The goal is to move notebook from the ground to the table, lay it on the table, and fold it. Each stage can be characterized by distinct contact mode. The positions and sizes of the tables and obstacles are randomized. The floating gripper grasps the notebook in the middle of its edge. The task is considered success if the pairwise distance to the goal object point cloud is below a threshold.
\begin{figure*}[h]
    \centering
    \includegraphics[width=1\linewidth]{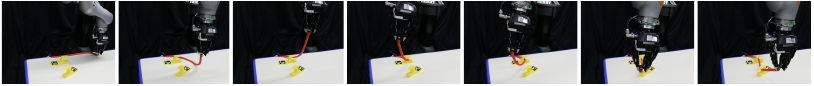}
    \caption{Physical demonstration with a 7-DoF Kuka arm on cable routing with 3 different cables for a total of 10 runs. Videos are available on our \href{https://implicit-contact-diffuser.github.io/}{website}.}
    \label{fig:physical}
    \vspace{-5mm}
\end{figure*}

For both tasks, the goal specification is provided as point cloud. We evaluated each method on 10 test cases for 2 seeds. 
The environments are built in the MuJoCo~\cite{todorov2012mujoco} simulator.
\subsection{Implementation Details}
We collected 5,000 trajectories of length 200 for cable routing and 10,000 trajectories of length 100 for notebook using scripted policies. The scene point clouds contains 1000 points and the object point clouds contain around 200 points. The NDF model is trained using equal weights for occupancy prediction and SDF gradient prediction. For VAE, the loss weights for reconstruction, KL-divergence and reachability are $1$, $1e^{-6}$ and $1e^{-5}$. For the diffusion model, we follow the training scheme of DDPMs~\cite{ho2020denoising} with 100 diffusion steps.
MPPI samples 80 trajectories with a horizon of 10. We use a noise scale of 0.001 for action sampling and a temperature of 0.005 for cost computation.

\subsubsection{Baselines}
1)~\textbf{MPPI}: We evaluate MPPI without the subgoals for guidance. We explore two different object representations for cost computation, referred to as PC-MPPI and NDF-MPPI; In PC-MPPI, the cost is computed as the distance in point cloud space, while NDF-MPPI computes cost in NDF space. 2)~\textbf{Subgoal Diffuser}~\cite{huang2024subgoal}: A modified version of Subgoal Diffuser that predicts a sequence of object point clouds using the same latent diffusion model as our method. The predicted subgoals are also tracked by the same MPPI planner. 3)~\textbf{Diffusion Policy}~\cite{chi2023diffusion}:  We adapt the official implementation to make the policy goal-conditioned. This version uses a keypoints-based object representation, while the scene information is encoded using the PointNet encoder from the NDF model.
4)~\textbf{3D Diffusion Policy}~\cite{ze20243d}: This baseline takes as input the point clouds of the object and the scene, and directly predicts the actions for the robot to execute.
5)~\textbf{Global NDF}. Instead of retrieving local features using KNN, this baseline follows the original NDF~\cite{simeonov2022neural} to compute a global feature vector for the entire scene.

\subsubsection{Results}
The quantitative results can be found in Table~\ref{table:sim_result}, and here we discuss our main findings. 

\noindent\textbf{Subgoal generation is critical for long-horizon reasoning}. We observe that the subgoal-based methods outperform both model-free methods that do not have explicit global reasoning (diffusion policy and 3D diffusion policy) and MPC methods that plan directly to the goal (PC-MPPI and NDF-MPPI). This result shows the importance of reasoning over the intermediate contact sequences explicitly.

\noindent\textbf{Contact-aware state representation is critical for long-horizon contact reasoning}. We observe that while subgoal diffuser performs well on notebook folding, its success rate drops significantly on cable routing. Upon inspection, we found that the primary failure mode is that the point cloud-based subgoal tends to lead the MPC to local minima since it does not capture the contact relationship.
For instance, it may lead to a state where the cable is spatially close to the goal configuration but is positioned incorrectly, such as being on the wrong side of the fixtures.
In contrast, our method leverages NDF to capture the geometric relationships between the rope and the fixtures. The NDF-based subgoals provide better guidance for the MPC to reach desired contact relationships. We also observe that the modified locally-conditioned NDF better captures the contact relationships compared to the global NDF.

\subsubsection{Adaptation test}

\begin{figure}[h]
    \centering
    \vspace{-2mm}
    \includegraphics[width=0.9\linewidth]{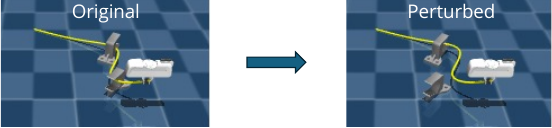}
        \caption{Examples for adaptation test.}
    \label{fig:adapt}
    \vspace{-3mm}
\end{figure}
Our previous experiment assumes the goal specification is provided for each task. However, obtaining the exact goal specification can be challenging in practice. It would be beneficial if we could reuse a previous goal specification $(\mP_{o_g}, P_s)$ in a different environment $P_s^\prime$. To explore this, we conduct an additional experiment, called the adaptation test. In this experiment, we randomly perturb the positions and orientations of the fixtures. As shown in Fig.~\ref{fig:adapt}, planning to the original goal using a point cloud-based representation is likely to fail, due to the change of fixture locations. Our key insight is that, even when the scene is altered, the desired contact relationships between the object and the scene remain consistent.

As illustrated in Table~\ref{table:adapt_results}, planning in the NDF space allows our method to successfully route the cable in the perturbed scene, whereas the baseline, which relies on precise positional subgoals, struggles to adapt to the changes in the environment.

\subsection{Physical Demonstration}
\begin{table}
\vspace{2mm}
    \centering
    \begin{tabular}{@{}c|c|c@{}}
        \toprule
        Method & Success rate $\uparrow$ & Complete Rate $\uparrow$  \\
        \midrule
        Ours &                                           90   & 90 \\
        Subgoal Diffuser~\cite{huang2024subgoal} &  25 & 47.5 \\
        \hline
    \end{tabular}
\caption{Results of adaptation test on cable routing}
\label{table:adapt_results}
\vspace{-6mm}
\end{table}

\begin{wrapfigure}{r}{0.2\textwidth}
    \centering
    \vspace{-4mm}
    \includegraphics[width=\linewidth]{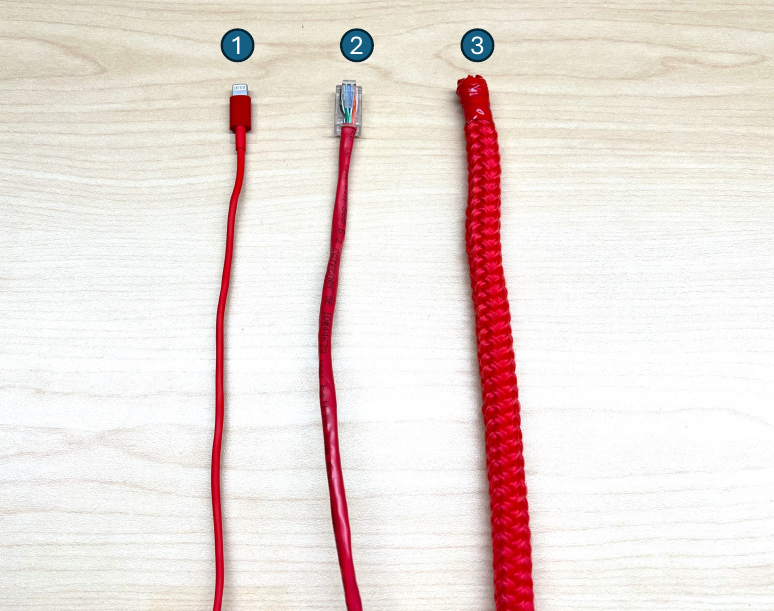}
    \caption{Cables used in the physical experiments.}
    \label{fig:real_world_rope}
    \vspace{-4mm}
\end{wrapfigure}
We deployed \method on a 7 DoF Kuka LBR iiwa arm for a real-world version of the cable-routing task. We used a Zivid 2 camera and CDCPD~\cite{wang2021tracking} to track the point cloud of the cable. We tested on 3 different cables, one soft, thin charging cable, one stiff ethernet cable, and a thick rope, for a total of 10 trials. 
While our method succeeds 9 / 10 runs, challenges such as perception errors from the tracker and the limited workspace of the robot affected the overall reliability of the method. 
Please see our \href{https://implicit-contact-diffuser.github.io/}{website} for the videos.

\section{Conclusion and Future Work}
We introduce a novel framework that enables the robot to reason about changing contacts between environments and  Our approach captures object-environment interactions using a smooth, continuous implicit representation. We then use a latent point cloud diffusion model to generate future contact subgoals using this representation. When integrated with an MPC method, the robot can intelligently initiate and break contacts to manipulate the object to satisfy a desired contact specification. However, the method has limitations: 1) It assumes access to full object and environment point clouds, which are often unavailable in real-world scenarios. Shape completion methods~\cite{chi2021garmentnets, huang2022mesh} could be applied to address this issue. 2) While replanning helps address model and perception errors, these errors are not considered during subgoal generation---a gap that could be addressed with online learning through interaction.

\newpage







\bibliographystyle{IEEEtran}
\bibliography{my}

\end{document}